\documentclass[a4paper, 10pt, conference]{ieeeconf}

\IEEEoverridecommandlockouts
\usepackage{amsfonts, amsmath, amssymb}
\usepackage{array, multirow}
\usepackage[english]{babel}
\usepackage[style=ieee, doi=false, url=false, isbn=false, mincitenames=1, maxcitenames=1, minbibnames=1, maxbibnames=1, backend=biber]{biblatex}
\usepackage{booktabs}
\usepackage[font=small]{caption, subcaption}
\usepackage[utf8]{inputenc}
\usepackage{csquotes}
\usepackage{float}
\usepackage[T1]{fontenc}
\usepackage[pdftex]{graphicx}
\usepackage{hyperref}
\usepackage[capitalize]{cleveref}
\usepackage{makecell}
\usepackage{pgfplots, pgfplotstable}

\graphicspath{{figures/}}
\hypersetup{colorlinks=true, linkcolor=black, citecolor=black, filecolor=black, urlcolor=black}
\AtBeginBibliography{\small}

\pgfplotsset{
    compat=newest,
    width=0.98\columnwidth,
    height=4cm,
    xmin=0,
    ymin=0,
    cycle list name=exotic,
    log plot exponent style/.style={/pgf/number format/precision=0},
    every axis/.append style={
        thick,
        tick style={thick}
    },
    legend style={line width=0.8pt}
}
\pgfplotstableset{col sep=comma}
\usepgfplotslibrary{groupplots}

\newcommand{\norm}[1]{\left\lVert#1\right\rVert}
\newcommand{\bel}[1]{\operatorname{bel(#1)}}

\title{Learned Enrichment of Top-View Grid Maps Improves Object Detection}

\author{Sascha Wirges$^{1}$, Ye Yang$^{1}$, Sven Richter$^{2}$, Haohao Hu$^{2}$ and Christoph Stiller$^{1,2}$
    \thanks{
        $^{1}$Authors are with Mobile Perception Systems Group,
        FZI Research Center for Information Technology,
        Karlsruhe, Germany --
        {\tt\footnotesize wirges@fzi.de}
    }
    \thanks{
        $^{2}$Authors are with Institute of Measurement and Control Systems,
        Karlsruhe Institute of Technology (KIT),
        Karlsruhe, Germany --
        {\tt\footnotesize \{sven.richter, haohao.hu, stiller\}@kit.edu}
    }
}

\begin{document}

\maketitle
\thispagestyle{empty}
\pagestyle{empty}

\begin{abstract}
	We propose an object detector for top-view grid maps which is additionally trained to generate an enriched version of its input.
	Our goal in the joint model is to improve generalization by regularizing towards structural knowledge in form of a map fused from multiple adjacent range sensor measurements.
	This training data can be generated in an automatic fashion, thus does not require manual annotations.
	We present an evidential framework to generate training data, investigate different model architectures and show that predicting enriched inputs as an additional task can improve object detection performance.
\end{abstract}
 \section{Introduction} \label{sec:introduction}

Automated driving requires environment models that provide information, e.g. about other traffic participants, at a high rate and precision.
However, environment models estimated from single measurements are often subject to noise and occlusions.
These disadvantages can be mitigated if multiple measurements from different viewpoints are considered in order to estimate an enriched map of the environment (see \cref{fig:transformation}).
In a post-processing step, this can be achieved by Simultaneous Localization and Mapping (SLAM) methods (e.g. \cite{Chen2019}) that fuse measurements in an acausal manner.

Given single range sensor measurements, enriched grid maps can be inferred by deep models trained on automatically generated \cite{Wirges2018Hartenbach} or on semantically annotated maps \cite{Yang2018} due to extensive offline processing.
Using these maps, Yang et al. show that estimating additional semantic information can increase object detection accuracy.

Here, we apply our methods to multi-layer top-view grid maps.
Due to the orthographic projection, observations are scale-invariant and do not overlap which makes grid maps well-suited for sensor fusion.
Their regular grid structure enables the use of efficient image processing operations such as convolutions or cell-wise operations.
Since all traffic participants move on a common ground surface, we believe a two-dimensional environment model along the ground surface is sufficient to represent the traffic environment.
Therefore, we assume objects standing on the ground surface and encode ground surface elevation and obstacle height as two layers.
Other layers may be the reflected energy or the evidences estimated during sensor fusion.

\begin{figure}[ht]
    \centering
    \scriptsize
    \def\svgwidth{\columnwidth}
    \begingroup
  \makeatletter
  \providecommand\color[2][]{
    \errmessage{(Inkscape) Color is used for the text in Inkscape, but the package 'color.sty' is not loaded}
    \renewcommand\color[2][]{}
  }
  \providecommand\transparent[1]{
    \errmessage{(Inkscape) Transparency is used (non-zero) for the text in Inkscape, but the package 'transparent.sty' is not loaded}
    \renewcommand\transparent[1]{}
  }
  \providecommand\rotatebox[2]{#2}
  \newcommand*\fsize{\dimexpr\f@size pt\relax}
  \newcommand*\lineheight[1]{\fontsize{\fsize}{#1\fsize}\selectfont}
  \ifx\svgwidth\undefined
    \setlength{\unitlength}{595.41820466bp}
    \ifx\svgscale\undefined
      \relax
    \else
      \setlength{\unitlength}{\unitlength * \real{\svgscale}}
    \fi
  \else
    \setlength{\unitlength}{\svgwidth}
  \fi
  \global\let\svgwidth\undefined
  \global\let\svgscale\undefined
  \makeatother
  \begin{picture}(1,0.30956405)
    \lineheight{1}
    \setlength\tabcolsep{0pt}
    \put(0,0){\includegraphics[width=\unitlength,page=1]{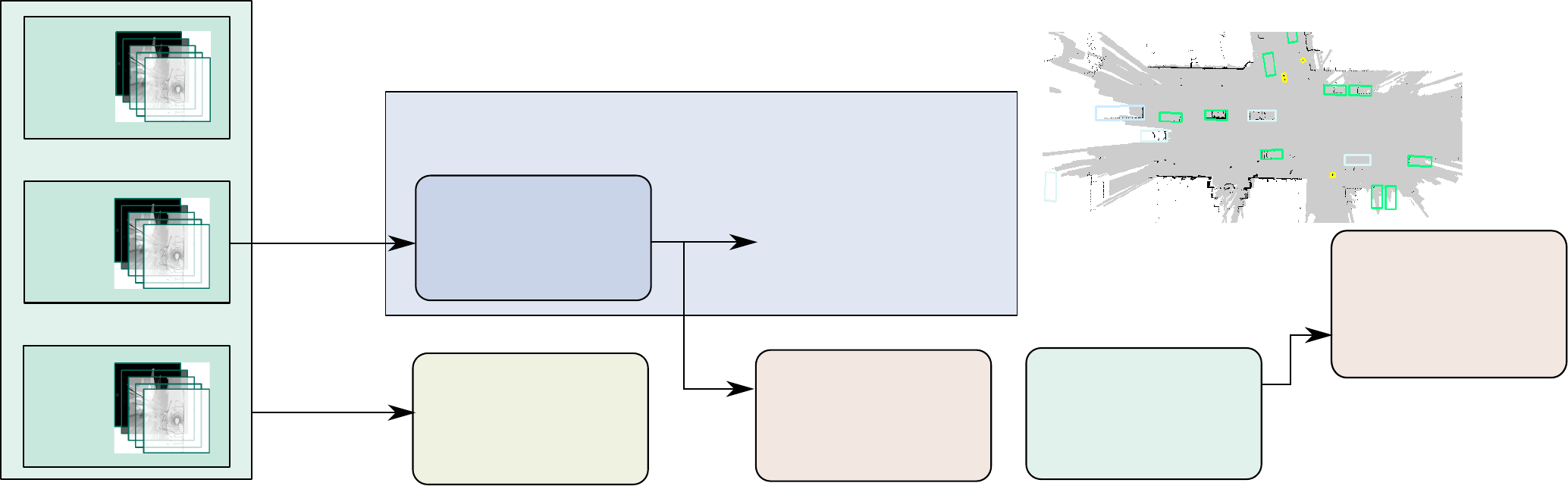}}
    \put(0.83462451,0.35424113){\color[rgb]{0,0,0}\makebox(0,0)[lt]{\begin{minipage}{0.04676097\unitlength}\centering \end{minipage}}}
    \put(3.16594661,0.34099151){\color[rgb]{0,0,0}\makebox(0,0)[lt]{\begin{minipage}{0.40576216\unitlength}\raggedright \end{minipage}}}
    \put(0,0){\includegraphics[width=\unitlength,page=2]{system_overview.pdf}}
    \put(0.02934035,0.0366422){\color[rgb]{0,0,0}\makebox(0,0)[lt]{\begin{minipage}{0.0454587\unitlength}\raggedright \end{minipage}}}
    \put(0.69109302,0.24996476){\color[rgb]{0,0,0}\makebox(0,0)[lt]{\begin{minipage}{0.46335983\unitlength}\raggedright \end{minipage}}}
    \put(0,0){\includegraphics[width=\unitlength,page=3]{system_overview.pdf}}
    \put(0.02574345,0.27182467){\color[rgb]{0,0,0}\makebox(0,0)[lt]{\begin{minipage}{0.0454587\unitlength}\raggedright t-k\end{minipage}}}
    \put(0.04086042,0.16644925){\color[rgb]{0,0,0}\makebox(0,0)[lt]{\begin{minipage}{0.0454587\unitlength}\raggedright t\end{minipage}}}
    \put(0.26739957,0.18404821){\color[rgb]{0,0,0}\makebox(0,0)[lt]{\begin{minipage}{0.15008584\unitlength}\centering Enrichment model\end{minipage}}}
    \put(0.26269303,0.0587195){\color[rgb]{0,0,0}\makebox(0,0)[lt]{\begin{minipage}{0.15008584\unitlength}\centering Fusion\end{minipage}}}
    \put(0,0){\includegraphics[width=\unitlength,page=4]{system_overview.pdf}}
    \put(0.48354695,0.1847714){\color[rgb]{0,0,0}\makebox(0,0)[lt]{\begin{minipage}{0.15008584\unitlength}\centering Object detector\end{minipage}}}
    \put(0.46876843,0.07173208){\color[rgb]{0,0,0}\makebox(0,0)[lt]{\begin{minipage}{0.1702897\unitlength}\centering Reconstr.\\ loss\end{minipage}}}
    \put(0.85258688,0.1574428){\color[rgb]{0,0,0}\makebox(0,0)[lt]{\begin{minipage}{0.15708407\unitlength}\centering Object detection\\ loss\end{minipage}}}
    \put(0.64393693,0.07459142){\color[rgb]{0,0,0}\makebox(0,0)[lt]{\begin{minipage}{0.17635493\unitlength}\centering Object\\ labels\end{minipage}}}
    \put(0.02173932,0.05575667){\color[rgb]{0,0,0}\makebox(0,0)[lt]{\begin{minipage}{0.06615244\unitlength}\raggedright t+k\end{minipage}}}
    \put(0.44679547,0.21990891){\color[rgb]{0,0,0}\makebox(0,0)[t]{\lineheight{1.25}\smash{\begin{tabular}[t]{c}Joint model\end{tabular}}}}
  \end{picture}
\endgroup
     \caption{
        Overview of the training procedure.
        Input for our joint model is a multi-layer grid map (time t).
        The reconstruction loss is determined based on the enriched grid maps and a fused map of all grid maps within the time horizon $\left[\mathrm{-k}, \mathrm{k}\right]$.
        The object detection loss is determined from estimated and labeled objects.
    } \label{fig:overview}
\end{figure}

\begin{figure}[ht]
    \begin{subfigure}{0.49\linewidth}
        \includegraphics[angle=-90, width=\linewidth]{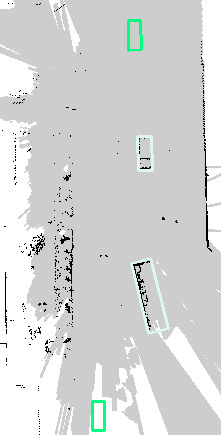}
        \caption{Baseline (single shot)} \label{fig:od_gridmap_detObs_baseline}
    \end{subfigure}
    \begin{subfigure}{0.49\linewidth}
        \includegraphics[angle=-90, width=\linewidth]{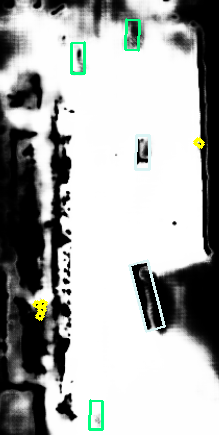}
        \caption{Shared encoder model} \label{fig:intro_gridmap_enriched_ev1}
    \end{subfigure}
    \caption{
        Our joint object detection and map enrichment model (\cref{fig:intro_gridmap_enriched_ev1}, detections with predicted free belief) yields improved performance, especially at larger distances or for smaller objects such as pedestrians, compared to a baseline trained without the enrichment task (\cref{fig:od_gridmap_detObs_baseline}, detections with observable space and reflections).
        Cars are depicted in green, trucks in white and pedestrians in yellow.
    } \label{fig:qualitative_results_front}
\end{figure}

In this work we study effects on object detection accuracy in presence of a second map enrichment task.
After briefly discussing related work on object detection, map enrichment and multi-task models in \cref{sec:related_work} we present our evidential framework to create top-view grid maps from single and multiple poses (\cref{sec:data_processing}).
We then propose sequential and shared encoder models for solving both enrichment and object detection tasks in \cref{sec:models_and_training}.
After providing a quantitative and qualitative evaluation of our models based on the nuScenes data set (\cref{sec:evaluation}), we conclude our work and point to future work in \cref{sec:conclusion}.
 \section{Related Work} \label{sec:related_work}

\subsection{Object Detection in Top-View Grid Maps} \label{sec:related_work_object_detection}
The performance of convolutional object detectors continuously increased in recent years.
Single-stage detectors (e.g. \cite{Liu2016}) have shown to be promising, as they produce accurate estimates at frame rates that are acceptable for real-time applications.
To mitigate the class-imbalance problem, Lin et al. \cite{Lin2017} propose a loss focusing on harder examples and employ a feature pyramid network (FPN) which reuses features at different pyramid levels.
Recent 3D object detection approaches (e.g. \cite{Qi2017, Lang2019}) use low-level encoders working on point sets that produce a feature representation which is then inserted into a volume grid structure.
In order to avoid expensive computation, Lang et al. \cite{Lang2019} reduce pillars of 3D features into 2D features.
Compared to high-dimensional feature representations, we use interpretable features for object detection \cite{Wirges2018Fischer} which are represented by different grid map layers.
That way we are able to combine a learned object detector with unsupervised clustering approaches in order to detect obstacles not represented in the labeled data set.
Usually we assume a static environment in occupancy grid mapping which is not satisfied in most scenarios.
Approaches to cope with moving objects include dynamic occupancy grid maps based on finite set statistics \cite{Nuss2018} or the detection and removal of dynamic parts of the scene \cite{Sahdev2017}.
Here, evidence theory allows to model contradicting measurements in order to remove or exclude uncertain areas \cite{Wirges2018Hartenbach}.

\subsection{Environment Enrichment} \label{sec:related_work_environment_enrichment}
Environment enrichment (sometimes augmentation) describes the process of inferring a complete environment model from observations subject to sensor noise and occlusions \cite{Wirges2018Hartenbach}.
Many approaches focus on reconstructing objects of interest (e.g. \cite{Menze2015, Mescheder2019}) with parametric models.
These methods either require ground-truth objects, do not consider the whole scenery or are computationally expensive.
Due to the unavailability of manually annotated data sets, self-supervised methods are common to train models for enrichment \cite{Wirges2018Hartenbach}.
The authors generate target data by fusing measurements in a 3D octree and use evidential combination rules along pillars to obtain an evidential 2D grid map representation.
Each cell of this representation then contains the free belief $\bel{F}$, occupied belief $\bel{O}$ and the belief
\begin{equation*}
	\bel{U} = 1 - \bel{F} - \bel{O}
\end{equation*}
for cells being of unknown state.
Every observed reflection or transmission obtained by casting rays from sensor origin to the reflection position contributes with elementary evidences $e_{R}(\{O\})$ and $e_{T}(\{F\})$, respectively.

Other approaches include estimating evidential 2D occupancy grid maps based on a sensor model parameterized with false positive probabilities $p_{\mathrm{FP}}$ and false negative probabilities $p_{\mathrm{FN}}$, respectively \cite{Richter2019}.
While they assign $p_{\mathrm{FP}}$ a constant value,
\begin{equation*}
	p_{\mathrm{FN}}\left(\boldsymbol{x}_{\mathrm{i}}\right)=1-\left(1-r_{\boldsymbol{x}_{\mathrm{i}}}\right) r_{z_{\mathrm{i}}}\left(1 - p_{\mathrm{FN, max}} \right)
\end{equation*}
depends on a maximum false positive probability $p_{\mathrm{FN, max}}$, the distance ratio
\begin{equation*}
	r_{\boldsymbol{x}_{\mathrm{i}}}=\frac{\norm{\boldsymbol{x}_{\mathrm{i}}}}{\norm{\boldsymbol{x}_{\max}}}
\end{equation*}
of measured distance $\norm{\boldsymbol{x}_{\mathrm{i}}}$ and maximum distance $\norm{\boldsymbol{x}_{\max}}$ and the height ratio
\begin{equation*}
	r_{z_{\mathrm{i}}}=\frac{\Delta z_{\mathrm{i}}}{\Delta z}
\end{equation*}
of observable height range $\Delta z_{\mathrm{i}}$ and relevant height range $\Delta z$.
Thus, $p_{\mathrm{FN}}$ increases in case of occlusions (reduced observable height) or large measurement distances.
Finally, the basic belief assignment (BBA)
\begin{equation}
	\mathrm{m}_{r}(\theta)=\left\{\begin{array}{ll}{p_{\mathrm{FN}}^{m}\left(1-p_{\mathrm{FP}}^{n}\right),} & {\text { if } \theta=\mathcal{O}} \\ {p_{\mathrm{FP}}^{n}\left(1-p_{\mathrm{FN}}^{m}\right),} & {\text { if } \theta=\mathcal{G}} \\ {1-\sum_{\xi \neq \Theta} m(\xi),} & {\text { if } \theta=\Theta}\end{array}\right.
	\label{eq:bba_richter}
\end{equation}
resembles belief in cells being obstacles ($\mathcal{O}$), ground reflections ($\mathcal{G}$) or unknown ($\Theta$), given the number of transmissions $m$ and reflections $n$.

\subsection{Joint Approaches} \label{sec:related_work_joint_approaches}
Yang et al. \cite{Yang2018} combine an object detector with a road mask prediction module in a multi-task network and show that average object detection precision can be improved.
They argue that additional information such as ground surface or road masks provide strong semantic information, especially at increasing distance.
The joint model can be realized either sequentially or as shared encoder architecture with a separate decoder for each task.
Learning auxiliary tasks can thus be regarded as a regularizer avoiding overfitting.

The final loss function in a multi-task setting usually resembles a linear combination of different task-specific loss functions.
Whereas search-based methods such as grid search to tune the task weights are computationally expensive, Kendall et al. interpret task-specific weights as uncertainties that are estimated during training for each task \cite{Kendall2018}.
 \section{Data Processing / Generation} \label{sec:data_processing}

\subsection{Data Set} \label{sec:data_processing_dataset}
The nuScenes object detection benchmark \cite{Caesar2019} is a public data set with multi-modal sensor data collected in Boston and Singapore traffic.
The data set consists of 1000 scenes of 20 seconds length.
It includes approximately 390k spinning range sensor sweeps in which 40k key frames are annotated with 23 classes of relevant objects such as cars, trucks, cyclists and pedestrians.
The spinning range sensor has 32 rays and operates at a frequency of 20Hz.
In addition, an accurate vehicle pose is provided which can be used to fuse range sensor measurements.

\subsection{Input Data} \label{sec:data_processing_input_data}
In the following we describe the grid mapping process in which single range sensor scans, represented by point sets, are mapped into measurement grid maps with a cell size of 15cm.
Given all points, we estimate the ground surface height by a cubic uniform b-spline in order to distinguish between ground and non-ground measurements.
In the first mapping step we determine for each grid cell the sum of non-ground reflections, the height and the average reflected energy.
By casting rays between sensor origin and reflection we obtain for each cell the number of transmissions, the height of cast shadows (assuming impenetrable obstacles) $z_\mathrm{i, shadow}$ and the maximum observable height $z_\mathrm{i, max. obs.}$.
We can then use the BBA in \cref{eq:bba_richter} with
\begin{equation*}
	\Delta z_\mathrm{i} = z_\mathrm{i, max. obs.} - z_\mathrm{i, shadow}
\end{equation*}
to determine the beliefs for cells being occupied or free.

\subsection{Target Data} \label{sec:data_processing_target_data}
\begin{figure}[ht]
	\centering
	\small
	\begin{subfigure}{0.45\linewidth}
		\def\svgwidth{\columnwidth}
		\begingroup
  \makeatletter
  \providecommand\color[2][]{
    \errmessage{(Inkscape) Color is used for the text in Inkscape, but the package 'color.sty' is not loaded}
    \renewcommand\color[2][]{}
  }
  \providecommand\transparent[1]{
    \errmessage{(Inkscape) Transparency is used (non-zero) for the text in Inkscape, but the package 'transparent.sty' is not loaded}
    \renewcommand\transparent[1]{}
  }
  \providecommand\rotatebox[2]{#2}
  \newcommand*\fsize{\dimexpr\f@size pt\relax}
  \newcommand*\lineheight[1]{\fontsize{\fsize}{#1\fsize}\selectfont}
  \ifx\svgwidth\undefined
    \setlength{\unitlength}{149.04502857bp}
    \ifx\svgscale\undefined
      \relax
    \else
      \setlength{\unitlength}{\unitlength * \real{\svgscale}}
    \fi
  \else
    \setlength{\unitlength}{\svgwidth}
  \fi
  \global\let\svgwidth\undefined
  \global\let\svgscale\undefined
  \makeatother
  \begin{picture}(1,0.46093612)
    \lineheight{1}
    \setlength\tabcolsep{0pt}
    \put(0,0){\includegraphics[width=\unitlength,page=1]{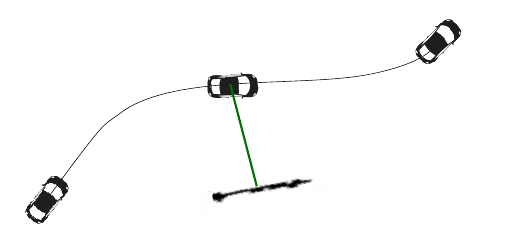}}
    \put(0.51861071,0.31470809){\color[rgb]{0,0.44313725,0}\makebox(0,0)[lt]{\begin{minipage}{0.15883668\unitlength}\raggedright t\end{minipage}}}
    \put(0.14065015,0.60773043){\color[rgb]{0,0,0}\makebox(0,0)[lt]{\begin{minipage}{0.33157525\unitlength}\raggedright \end{minipage}}}
    \put(0.90475787,0.42246319){\color[rgb]{0.66666667,0.83137255,0}\makebox(0,0)[lt]{\begin{minipage}{0.15883668\unitlength}\raggedright t+1\end{minipage}}}
    \put(0.14384642,0.14758538){\color[rgb]{0.26666667,0.25490196,0}\makebox(0,0)[lt]{\begin{minipage}{0.15883668\unitlength}\raggedright t-1\end{minipage}}}
    \put(-0.27629,0.25638646){\color[rgb]{0,0,0}\makebox(0,0)[lt]{\begin{minipage}{0.42323018\unitlength}\raggedright \end{minipage}}}
    \put(0,0){\includegraphics[width=\unitlength,page=2]{transformations_single.pdf}}
  \end{picture}
\endgroup
 		\caption{Single scan} \label{fig:transformation_single_scan}
	\end{subfigure}
	\begin{subfigure}{0.45\linewidth}
		\def\svgwidth{\columnwidth}
		\begingroup
  \makeatletter
  \providecommand\color[2][]{
    \errmessage{(Inkscape) Color is used for the text in Inkscape, but the package 'color.sty' is not loaded}
    \renewcommand\color[2][]{}
  }
  \providecommand\transparent[1]{
    \errmessage{(Inkscape) Transparency is used (non-zero) for the text in Inkscape, but the package 'transparent.sty' is not loaded}
    \renewcommand\transparent[1]{}
  }
  \providecommand\rotatebox[2]{#2}
  \newcommand*\fsize{\dimexpr\f@size pt\relax}
  \newcommand*\lineheight[1]{\fontsize{\fsize}{#1\fsize}\selectfont}
  \ifx\svgwidth\undefined
    \setlength{\unitlength}{149.12599904bp}
    \ifx\svgscale\undefined
      \relax
    \else
      \setlength{\unitlength}{\unitlength * \real{\svgscale}}
    \fi
  \else
    \setlength{\unitlength}{\svgwidth}
  \fi
  \global\let\svgwidth\undefined
  \global\let\svgscale\undefined
  \makeatother
  \begin{picture}(1,0.47262793)
    \lineheight{1}
    \setlength\tabcolsep{0pt}
    \put(0,0){\includegraphics[width=\unitlength,page=1]{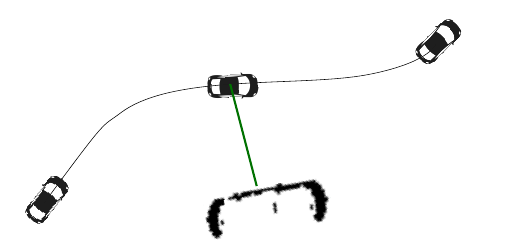}}
    \put(0.51832911,0.32647929){\color[rgb]{0,0.44313725,0}\makebox(0,0)[lt]{\begin{minipage}{0.15875044\unitlength}\raggedright t\end{minipage}}}
    \put(0.90426663,0.43417589){\color[rgb]{0.66666667,0.83137255,0}\makebox(0,0)[lt]{\begin{minipage}{0.15875044\unitlength}\raggedright t+1\end{minipage}}}
    \put(0.14376837,0.15944733){\color[rgb]{0.26666667,0.25490196,0}\makebox(0,0)[lt]{\begin{minipage}{0.15875044\unitlength}\raggedright t-1\end{minipage}}}
    \put(0,0){\includegraphics[width=\unitlength,page=2]{transformations_subsequent.pdf}}
  \end{picture}
\endgroup
 		\caption{Three scans} \label{fig:transformation_three_scans}
	\end{subfigure}
	\caption{
		As depicted in \cref{fig:transformation_single_scan}, objects are usually only observed partly from single viewpoints, e.g. due to occlusions.
		Fusing observations from different viewpoints during driving as depicted in \cref{fig:transformation_three_scans} results in a more accurate object reconstruction.
	} \label{fig:transformation}
\end{figure}

As illustrated in \cref{fig:transformation} the target data should consist of a complete scene fused from measurements at different viewpoints.
Therefore, we fuse data within a defined radius of 40m around a reference pose and resolve contradictions due to moving traffic participants using evidential combination rules assuming measurements from different time steps.
In contrast to \cite{Wirges2018Hartenbach}, we directly fuse measurement grid maps in this work.
Using this method, we did not observe any accuracy degradation at the advantage of parallelization and thus faster map generation.

Within a time interval $\mathcal{T}$ with $N$ observations, we model the frame of discernment
\begin{equation*}
	\Omega = \{ \underbrace{\left( \omega_1, \ldots, \omega_N \right)}_{\boldsymbol{\omega}}~|~\omega_i \in \{ \mathrm{o}, \mathrm{f} \} \}
\end{equation*}
as the set of tuples of individual cell states corresponding to time points $t_1 \ldots, t_N \in \mathcal{T}$.
By assuming temporally independent cell states, we define the basic belief assignment as
\begin{equation*}
	\mathrm{m}(A) = \prod\limits_{i = 1}^N \mathrm{m}_r\left(\bigcup_{\omega\in A} \{\omega_i\}\right)
\end{equation*}
for $A \in 2^\Omega$.

\begin{figure}[ht]
	\centering
	\begin{subfigure}{0.49\linewidth}
		\includegraphics[angle=-90, width=\linewidth]{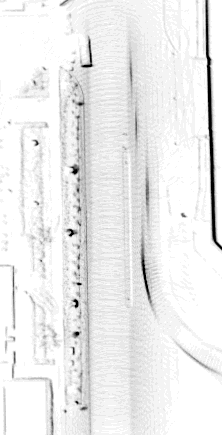}
		\caption{Reflections} \label{fig:gridmap_fused_reflections}
	\end{subfigure}
	\begin{subfigure}{0.49\linewidth}
		\includegraphics[angle=-90, width=\linewidth]{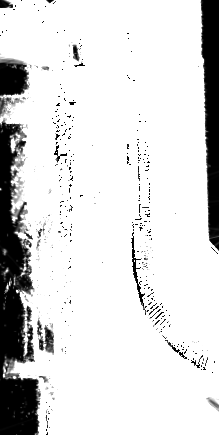}
		\caption{$\bel{\mathrm{U}}$} \label{fig:gridmap_belU}
	\end{subfigure}
	\caption{
		Target reflection and uncertainty belief.
		While the fused reflections contain artifacts due to moving objects, grid cells corresponding to these objects are assigned to uncertainty.
	} \label{fig:overview_fused_data}
\end{figure}

We introduce the partition
\begin{equation*}
	2^\Omega = \mathcal{F} \,\dot\cup\, \mathcal{O} \,\dot\cup\, \mathcal{D},
\end{equation*}
where
\begin{align*}
	\mathcal{F} & = \{ \boldsymbol{\omega}~|~\omega_i \in \{ \mathrm{f}, \{ \mathrm{o}, \mathrm{f} \} \},    \exists i : \omega_i = \mathrm{f} \}, \\
	\mathcal{O} & = \{ \boldsymbol{\omega}~|~\omega_i \in \{ \mathrm{o}, \{ \mathrm{o}, \mathrm{f} \} \},    \exists i : \omega_i = \mathrm{o} \}, \\
	\mathcal{D} & = 2^\Omega \setminus \{ \mathcal{F} \cup \mathcal{O} \}.
\end{align*}
The hypotheses in $\mathcal{F}$ describe that at least one cell was observed as free and none as occupied, the ones in $\mathcal{O}$ that at least one cell was observed as occupied but none as free.
The remaining hypotheses in $\mathcal{D}$ represent all observation sequences were a cell was observed at least once free \textit{and} at least once occupied indicating the dynamic parts of the scene.
Consequently, we model the fused masses for occupied and free as
\begin{equation*}
	\bel{\mathcal{X}} = \sum\limits_{\boldsymbol{\omega}\subset\mathcal{X}} m(\boldsymbol{\omega}),\quad \mathcal{X}\in\{\mathcal{O}, \mathcal{F}\},
\end{equation*}
i.e. explicitly ignoring dynamic parts of the scene (see \cref{fig:overview_fused_data}).

The height of obstacles above ground is modeled by independent normal distributions $\mathrm{p}\left(z | \mu_i, \sigma_i\right)$ with mean
\begin{equation*}
	\mu_i = \frac{z_{i, \mathrm{max. obs.}} + z_{i, \mathrm{max. det.}}}{2}
\end{equation*}
and variance
\begin{equation*}
	\sigma_i^2 = z_{i, \mathrm{max. obs.}} - z_{i, \mathrm{max. det.}}
\end{equation*}
which enables us to estimate the height distribution
\begin{equation*}
	\mathrm{p}\left(z | \mu_1, \sigma_1, \mu_2, \sigma_2, \ldots, \mu_N, \sigma_N\right) = \prod_{i=1}^N \mathrm{p}\left(z | \mu_i, \sigma_i\right)
\end{equation*}
\Cref{tab:used_layers} summarizes single frame and fused layers used in this work.

\begin{table}[ht]
	\centering
	\begin{tabular}{l|l}
		\textbf{Input layers (single frame)}                               & \textbf{Target layers (fused)}                     \\ \hline
		Reflections (black cells in \cref{fig:od_gridmap_detObs_baseline}) & Reflections (\cref{fig:gridmap_fused_reflections}) \\
		Observations (grey cells in \cref{fig:od_gridmap_detObs_baseline}) & Observation height                                 \\
		Reflected energy                                                   & Reflected energy                                   \\
		Height (\cref{fig:gridmap_baseline_height})                        & Height                                             \\
		Height of cast shadows                                             & Free belief                                        \\
		                                                                   & Occupied belief                                    \\
		                                                                   & Uncertainty belief (\cref{fig:gridmap_belU})       \\
	\end{tabular}
	\caption{
		Grid map layers used in the baseline and joint models.
		Height layers are relative to the estimated ground surface.
	} \label{tab:used_layers}
\end{table}
 \section{Models and Training Procedure} \label{sec:models_and_training}

We first introduce our baseline object detector and then present a sequential and a shared-encoder structure for joint object detection and grid map enrichment.

\begin{figure*}[ht]
	\centering
	\small
	\begin{subfigure}{0.49\linewidth}
		\def\svgwidth{\columnwidth}
		\begingroup
  \makeatletter
  \providecommand\color[2][]{
    \errmessage{(Inkscape) Color is used for the text in Inkscape, but the package 'color.sty' is not loaded}
    \renewcommand\color[2][]{}
  }
  \providecommand\transparent[1]{
    \errmessage{(Inkscape) Transparency is used (non-zero) for the text in Inkscape, but the package 'transparent.sty' is not loaded}
    \renewcommand\transparent[1]{}
  }
  \providecommand\rotatebox[2]{#2}
  \newcommand*\fsize{\dimexpr\f@size pt\relax}
  \newcommand*\lineheight[1]{\fontsize{\fsize}{#1\fsize}\selectfont}
  \ifx\svgwidth\undefined
    \setlength{\unitlength}{140.21058296bp}
    \ifx\svgscale\undefined
      \relax
    \else
      \setlength{\unitlength}{\unitlength * \real{\svgscale}}
    \fi
  \else
    \setlength{\unitlength}{\svgwidth}
  \fi
  \global\let\svgwidth\undefined
  \global\let\svgscale\undefined
  \makeatother
  \begin{picture}(1,0.65334568)
    \lineheight{1}
    \setlength\tabcolsep{0pt}
    \put(0,0){\includegraphics[width=\unitlength,page=1]{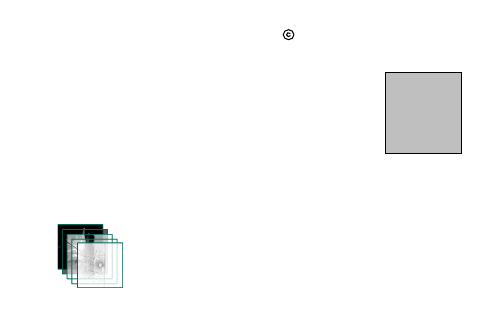}}
    \put(1.47485201,1.06053566){\color[rgb]{0,0,0}\makebox(0,0)[lt]{\begin{minipage}{0.46465742\unitlength}\raggedright \end{minipage}}}
    \put(1.95153935,0.83676322){\color[rgb]{0,0,0}\makebox(0,0)[lt]{\begin{minipage}{0.27350098\unitlength}\raggedright \end{minipage}}}
    \put(0.8705846,0.47462682){\color[rgb]{0,0,0}\makebox(0,0)[t]{\lineheight{1.25}\smash{\begin{tabular}[t]{c} \end{tabular}}}}
    \put(0,0){\includegraphics[width=\unitlength,page=2]{seq_model.pdf}}
    \put(-0.10145431,1.07523996){\color[rgb]{0,0,0}\makebox(0,0)[lt]{\begin{minipage}{1.538075\unitlength}\raggedright \end{minipage}}}
    \put(-0.28084738,1.1105304){\color[rgb]{0,0,0}\makebox(0,0)[lt]{\begin{minipage}{1.60571508\unitlength}\raggedright \end{minipage}}}
    \put(0.87011724,0.45479025){\color[rgb]{0,0,0}\makebox(0,0)[t]{\lineheight{1.25}\smash{\begin{tabular}[t]{c}Object\\detection\\network\end{tabular}}}}
    \put(0.34353613,0.2957892){\color[rgb]{0,0,0}\makebox(0,0)[lt]{\lineheight{1.25}\smash{\begin{tabular}[t]{l}Att.\\Gate\end{tabular}}}}
    \put(0.35345023,0.17700658){\color[rgb]{0,0,0}\makebox(0,0)[t]{\lineheight{1.25}\smash{\begin{tabular}[t]{c}Single-frame\\multi-layer\\grid map\end{tabular}}}}
    \put(0.5523639,0.07212923){\color[rgb]{0,0,0}\makebox(0,0)[lt]{\lineheight{1.25}\smash{\begin{tabular}[t]{l}Enriched layers\\and evidences\end{tabular}}}}
    \put(0,0){\includegraphics[width=\unitlength,page=3]{seq_model.pdf}}
  \end{picture}
\endgroup
 		\caption{Sequential models} \label{fig:overview_seq}
	\end{subfigure}
	\begin{subfigure}{0.49\linewidth}
		\def\svgwidth{\columnwidth}
		\begingroup
  \makeatletter
  \providecommand\color[2][]{
    \errmessage{(Inkscape) Color is used for the text in Inkscape, but the package 'color.sty' is not loaded}
    \renewcommand\color[2][]{}
  }
  \providecommand\transparent[1]{
    \errmessage{(Inkscape) Transparency is used (non-zero) for the text in Inkscape, but the package 'transparent.sty' is not loaded}
    \renewcommand\transparent[1]{}
  }
  \providecommand\rotatebox[2]{#2}
  \newcommand*\fsize{\dimexpr\f@size pt\relax}
  \newcommand*\lineheight[1]{\fontsize{\fsize}{#1\fsize}\selectfont}
  \ifx\svgwidth\undefined
    \setlength{\unitlength}{213.77912503bp}
    \ifx\svgscale\undefined
      \relax
    \else
      \setlength{\unitlength}{\unitlength * \real{\svgscale}}
    \fi
  \else
    \setlength{\unitlength}{\svgwidth}
  \fi
  \global\let\svgwidth\undefined
  \global\let\svgscale\undefined
  \makeatother
  \begin{picture}(1,0.66140824)
    \lineheight{1}
    \setlength\tabcolsep{0pt}
    \put(0,0){\includegraphics[width=\unitlength,page=1]{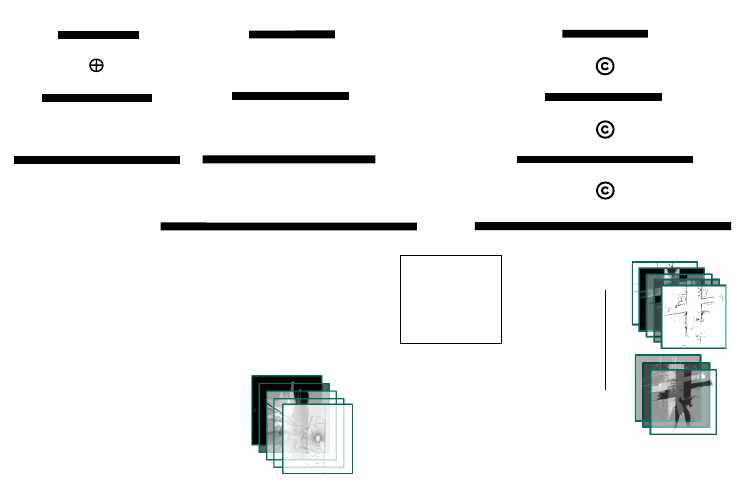}}
    \put(0.54688788,0.2933597){\color[rgb]{0,0,0}\makebox(0,0)[lt]{\begin{minipage}{0.08085521\unitlength}\raggedright \end{minipage}}}
    \put(0.56640273,0.26869199){\color[rgb]{0,0,0}\makebox(0,0)[lt]{\lineheight{1.25}\smash{\begin{tabular}[t]{l}Att.\\Gate\end{tabular}}}}
    \put(0.26165639,0.29971846){\color[rgb]{0,0,0}\makebox(0,0)[t]{\lineheight{1.25}\smash{\begin{tabular}[t]{c}Object\\detection\\network\end{tabular}}}}
    \put(0.58165803,0.11609272){\color[rgb]{0,0,0}\makebox(0,0)[t]{\lineheight{1.25}\smash{\begin{tabular}[t]{c}Single-frame\\multi-layer\\grid map\end{tabular}}}}
    \put(0.7121189,0.04730715){\color[rgb]{0,0,0}\makebox(0,0)[lt]{\lineheight{1.25}\smash{\begin{tabular}[t]{l}Enriched layers\\and evidences\end{tabular}}}}
    \put(0,0){\includegraphics[width=\unitlength,page=2]{sharedEncoder.pdf}}
  \end{picture}
\endgroup
 		\caption{Shared encoder models} \label{fig:overview_se}
	\end{subfigure}
	\caption{
		Evaluated sequential and shared encoder models (depicted here with attention gate).
		The sequential models (\cref{fig:overview_seq}) employ a separate modified MultiResUNet \cite{Ibtehaz2019} to infer enriched grid map layers and concatenate them with the single-frame inputs together to feed them into the detector.
		The shared-encoder models (\cref{fig:overview_se}) use a common backbone and two separate decoder branches.
		The enrichment branch also employs MultiRes blocks with optional self-attention mechanism \cite{Oktay2018}.
	} \label{fig:meta_arch_compare}
\end{figure*}

\subsection{Object Detection Network} \label{sec:sxod}
The architecture of our single-stage object detector with Feature Pyramid Network (FPN) is depicted in the gray box of \cref{fig:overview_se}.
As a trade-off between speed and accuracy, we use level 1-4 of the FPN without max-pooling in the root block and a modified ResNet-50 with the initial filter sizes reduced to 32, 64, 96 and 128, respectively.
The extracted feature maps in the bottom-up path are reshaped and added to those in the top-down path of the FPN before they are fed into the weight-shared box predictor.

\subsection{Enrichment Head and Loss Function} \label{sec:models_prediction_head}
The enrichment network head consists of two branches for inference of five grid map layers and three evidential maps, respectively.
While the branch for evidential maps subject to $\bel{O}+\bel{F}+\bel{\Theta} = 1$ is followed by a softmax activation yielding the loss $L_{\mathrm{ev}}$, ReLUs and the L1 loss $L_{\mathrm{gm}}$ are used for the second regression branch which is scaled by the mask
\begin{equation*}
	W_{k}=1-k \cdot \bel{U} \quad k \in \left[0,1\right]
\end{equation*}
depending on the target data uncertainty $\bel{U}$.
This scaling suppresses loss due to sensor noise, moving obstacles and the unobservable areas.
The parameter $k$ can be adjusted for each layer but is set to 0.9 in this work.

We employ task-uncertainty weighting for multi-task learning in order to balance the loss
\begin{equation*}
	\mathcal{L}_{\mathrm{enr}}
	\approx \frac{1}{2 \sigma_{1}^{2}} \mathcal{L}_{\mathrm{gm}}(W_k)+\frac{1}{\sigma_{2}^{2}} \mathcal{L}_{\mathrm{ev}}+\log \sigma_{1}+\log \sigma_{2}
\end{equation*}
for enrichment as well as the localization loss $L_{\mathrm{loc}}$ and the focal loss for classification $L_{\mathrm{cls}}$.
This results in the final loss
\begin{equation*}
	\mathcal{L} \approx \mathcal{L}_{\mathrm{enr}} + \frac{1}{2\sigma_{3}^{2}} \mathcal{L}_{\mathrm{loc}} + \frac{1}{\sigma_{4}^{2}} \mathcal{L}_{\mathrm{cls}} + \log \sigma_{3} + \log \sigma_{4}.
\end{equation*}

\subsection{Sequential Models (Seq-12 / Seq-32)} \label{sec:models_sequential}
\paragraph{UNet with MultiRes Block}
We employ a contraction-expansion structure with 4-fold successive downsampling, each followed by a MultiRes block \cite{Ibtehaz2019}.
Each block passes the input to one, two and three stacked convolutional layers and concatenates outputs before reducing it to a proper depth via 1x1 convolutions.
Similar to ResNet, the resulting feature maps are then added to their inputs.
The filter size starts from 12 or 32 and is doubled in each block after pooling.
Upsampled features in the expansion pathway are concatenated with the lateral short cuts.

\paragraph{Self-Attention Gate}
The self-attention gate \cite{Oktay2018} downsamples the lateral shortcut and concatenates it with the feature maps from the expansion pathway in order to make use of context and the local information.
This merged feature map is passed to a point-wise convolutional layer followed by sigmoid activation in order to generate the attention mask.
As the learned attention mask is applied to the non-downsampled lateral shortcut, we resize the weighting mask using bilinear interpolation.

\subsection{Shared Encoder Models (SE)} \label{sec:models_shared_encoder}
To reduce parameters and latency we develop a shared-encoder model with two decoders sharing generalized features provided by a modified ResNet-50 mentioned in \cref{sec:sxod}.
As shown in \cref{fig:overview_se}, the decoder for enrichment is the same as the one used in our sequential models.
The features for object detection are concatenated with features in the enrichment decoder before they are fed to the final detection head (not shown in \cref{fig:overview_se}).

\subsection{Training Details} \label{sec:models_training_details}

\begin{table}[ht]
    \begin{tabular}{c|c|c|c|c|c}
        \toprule
        \textbf{Configuration} & \textbf{Baseline} & \textbf{Target} & \textbf{Seq-12} & \textbf{Seq-32} & \textbf{SE} \\
        \midrule
        Add. conv. layers      & 0                 & 0               & 29              & 29              & 24          \\
        Epochs                 & 13                & 13              & 13              & 17              & 23          \\
        \bottomrule
    \end{tabular}
    \caption{
        Additional convolutional layers and epochs trained for evaluated model architectures.
    } \label{tab:trainingdetails}
\end{table}

All our models are trained using the Adam optimizer with a learning rate of $10^{-5}$.
For each model, we stopped the training when the validation error converged without overfitting.
Experiment details are summarized in \cref{tab:trainingdetails}.
The input data is preprocessed using random flips and fed into the network at a batch size of 2 due to memory limitations.
For the experiments \textit{Baseline} and \textit{Target} we use the same SSD subnet as described in \cref{sec:sxod}.
The difference here is that for \textit{Target} all grid map layers are used as input.
We use an NVIDIA GeForce RTX 2080 GPU for training and evaluation of all models. \section{Evaluation} \label{sec:evaluation}

\subsection{Grid Map Enrichment} \label{sec:evaluation_grid_map_enrichment}

\begin{table*}[ht]
    \begin{tabular*}{\linewidth}{@{\extracolsep{\fill}}c|c|c@{}c@{}c@{}c|c@{}c@{}c@{}c@{}c}
        \toprule
        \multirowcell{2}{\textbf{Model}\\ \textbf{Architecture}} & \multirowcell{2}{\textbf{Time} \\ \textbf{(ms)}} & \multicolumn{4}{c}{\textbf{Evidential Maps}} & \multicolumn{5}{c}{\textbf{Grid Map Layers (L1-Norm)}} \\ \cmidrule{3-11}
        &                                     &
        \textbf{{\begin{tabular}
                        {@{}c@{}}L1- \\Norm
                    \end{tabular}}}

        & \textbf{{\begin{tabular}
                        {@{}c@{}}L2- \\Norm
                    \end{tabular}} }

        & \textbf{{\begin{tabular}
                        {@{}c@{}}False \\Occup.
                    \end{tabular}}
        }
        & \textbf{
            {\begin{tabular}
                        {@{}c@{}}False \\Free
                    \end{tabular}}
        } & \textbf{Det.} & \textbf{Int.} & \textbf{{\begin{tabular}
                        {@{}c@{}}Z-max \\Det.
                    \end{tabular}}
        } & \textbf{{\begin{tabular}
                        {@{}c@{}}Z-min \\Det.
                    \end{tabular}}
        } & \textbf{{\begin{tabular}
                        {@{}c@{}}Z-min \\Obs.
                    \end{tabular}}
        } \\
        \midrule
        Seq-12 & \textbf{84} & 0.228 & 0.18 & 0.114 & 0.050 & 8.16 & 9.24 & 69.96 & 26.30 & 99.99\\
        Seq-32 & 98 & \textbf{0.113} & \textbf{0.076} & \textbf{0.042} & \textbf{0.029} & \textbf{2.52} & \textbf{3.61} & 30.63 & \textbf{10.09} & \textbf{42.08}\\
        Seq-32 (no att.) & 98 & 0.185 & 0.161 & 0.102 & 0.042 & 3.96 &  5.05 & 37.94 &  24.21 & 73.00 \\
        SE & 89 & 0.158 & 0.116 & 0.068 & 0.034 & 3.47 & 5.09 & \textbf{28.70} & 38.91 & 82.60\\
        \bottomrule
    \end{tabular*}
    \caption{
        Quantitative evaluation of different architectures regarding their augmentation performance.
        The best results are denoted in bold font.
    } \label{tab:aug_quanti}
\end{table*}

Similar to the evaluation in \cite{Wirges2018Hartenbach} we determine the cell-wise L1- and L2-norm of the difference between target and estimation.
Also, we determine the false occupied / free metrics
\begin{align*}
    m_{\mathrm{False O}} & = \max \left(0, \mathrm{bel}^{\prime}(O) + \bel{F} - 1 \right)       \\
    m_{\mathrm{False F}} & = \max \left(0, \bel{O} + \operatorname{bel}^{\prime}(F) - 1 \right)
\end{align*}
which penalize high estimated belief $\operatorname{bel}^{\prime}(\cdot)$ in contradiction to target data $\bel{\cdot}$.
\Cref{tab:aug_quanti} summarizes the results for different model configurations.
Here, the sequential model \textit{Seq-32} with attention gate yields the best performance with the shared encoder model \textit{SE} showing slightly worse performance.

\begin{figure}[ht]
    \centering
    \begin{subfigure}{0.49\linewidth}
        \includegraphics[angle=-90, width = \linewidth ]{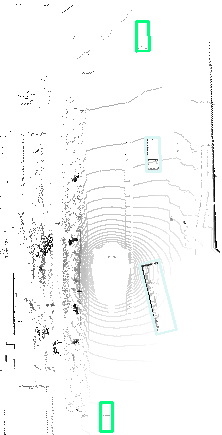}
        \caption{Baseline (height)} \label{fig:gridmap_baseline_height}
    \end{subfigure}
    \begin{subfigure}{0.49\linewidth}
        \includegraphics[angle=-90, width = \linewidth ]{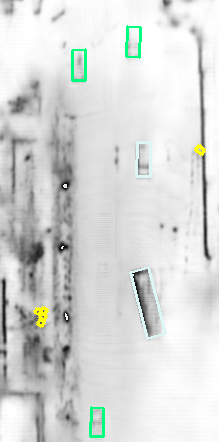}
        \caption{Shared encoder (height)} \label{fig:gridmap_enriched_height}
    \end{subfigure}

    \begin{subfigure}{0.49\linewidth}
        \includegraphics[angle=-90, width = \linewidth ]{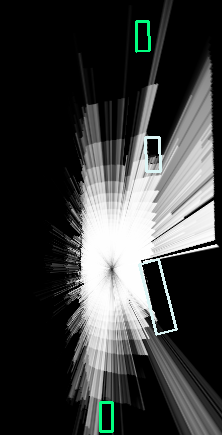}
        \caption{Baseline ($\bel{F}$)} \label{fig:gridmap_baseline_belF}
    \end{subfigure}
    \begin{subfigure}{0.49\linewidth}
        \includegraphics[angle=-90, width = \linewidth ]{n015-2018-08-01-16-41-59+0800__LIDAR_TOP__1533113037547164_bel_F_se}
        \caption{Shared encoder ($\bel{F}$)} \label{fig:gridmap_enriched_belF}
    \end{subfigure}

    \begin{subfigure}{0.49\linewidth}
        \includegraphics[angle=-90, width = \linewidth ]{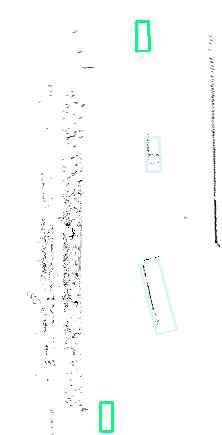}
        \caption{Baseline ($\bel{O}$)} \label{fig:gridmap_baseline_belO}
    \end{subfigure}
    \begin{subfigure}{0.49\linewidth}
        \includegraphics[angle=-90, width = \linewidth ]{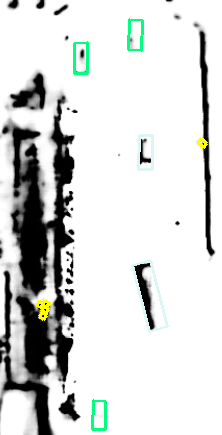}
        \caption{Shared encoder ($\bel{O}$)} \label{fig:gridmap_enriched_belO}
    \end{subfigure}
    \caption{
        Qualitative comparison of object detection performance between baseline and shared encoder model.
        We also denote enriched grid map layers in comparison to their corresponding single frame layers.
    } \label{fig:qualitative_results}
\end{figure}
\Cref{fig:qualitative_results} compares the enriched layer to its single frame layers for one particular scenario.
We observe that the height is estimated correctly for most of the cells and that information is visible in the enriched layers which is we can not recognize in the single frame layers.

\subsection{Object Detection} \label{sec:evaluation_object_detection}

\begin{table*}[ht]
    \resizebox{\linewidth}{!}{
        \begin{tabular*}{\linewidth}{@{
                    \extracolsep{\fill}}c|c@{}c@{}c@{}c@{}c|c@{}c@{}c@{}c|c|c|c|c}
            \toprule
            \multirowcell{2}{\textbf{Model}\\ \textbf{Archit.}}

            & \multicolumn{5}{c}{\textbf{AP of large objects}}
            & \multicolumn{4}{c}{\textbf{AP of small objects}}
            & \multirowcell{2}{\textbf{Mean} \\ \textbf{AP}}
            & \multirowcell{2}{\textbf{Mean} \\ \textbf{ATE}}
            & \multirowcell{2}{\textbf{Mean} \\ \textbf{ASE}}
            & \multirowcell{2}{\textbf{ND} \\ \textbf{Score}}
            \\ \cmidrule{2-10}
            & \textbf{Car}
            & \textbf{Truck}
            & \textbf{Bus}
            & \textbf{Trailer}
            & \textbf{{\begin{tabular}
                            {@{}c@{}}Cons. \\veh.
                        \end{tabular}}}
            & \textbf{Ped.}
            & \textbf{{\begin{tabular}
                            {@{}c@{}}Motor- \\cycle
                        \end{tabular}}}
            & \textbf{{\begin{tabular}
                            {@{}c@{}}Bi- \\cycle
                        \end{tabular}}}
            & \textbf{Barrier}
            &
            &
            & \\
            \midrule

            Target & 0.659 & 0.302 & 0.287 & 0.128 & 0.049 & 0.332 & 0.179 & 0.035  & 0.203& 0.218 & 0.437 & 0.616 & 0.218 \\
            Baseline & 0.593 & 0.220 & 0.289 & 0.052 & 0.009 & 0.162 & 0.108 & 0.001 & 0.192& 0.163 & 0.572 & 0.628 & 0.161 \\
            \midrule
            Seq-12 & 0.561 & 0.139 & \textbf{0.308} & 0.129 & 0.009 & 0.164 & 0.105 & 0.015  & \textbf{0.236} & 0.167  & \textbf{0.466} & \textbf{0.589} &  0.178 \\
            Seq-32 & 0.580 & 0.177 & 0.304 & 0.122 & 0.008 & 0.189 & 0.160 & 0.011 & 0.171 & 0.172  & \textbf{0.466} & 0.603 & 0.179 \\
            Seq-32 (no att.) & 0.568 & 0.188 & 0.294 & 0.102 &  0.012 & 0.245 &  0.145 & 0.006 &  0.141 & 0.170 & 0.472 &0.594 &0.178\\
            SE & 0.569 & 0.174 & \textbf{0.308} & \textbf{0.151} & 0.001 & \textbf{0.231} & \textbf{0.198} & \textbf{0.020} &  0.182 & \textbf{0.183} & 0.473 & 0.637 & \textbf{0.180} \\
            \bottomrule
        \end{tabular*}}
    \caption{
        Quantitative evaluation of different configurations regarding their object detection performance.
        The best model results that are better than the baseline are denoted by bold fonts.
    } \label{tab:od_quanti}
\end{table*}

\pgfplotstableread{results/mAP_large.csv}{\mAPLarge}
\pgfplotstableread{results/mAP_small.csv}{\mAPSmall}
\pgfplotstableread{results/ATE.csv}{\ATE}

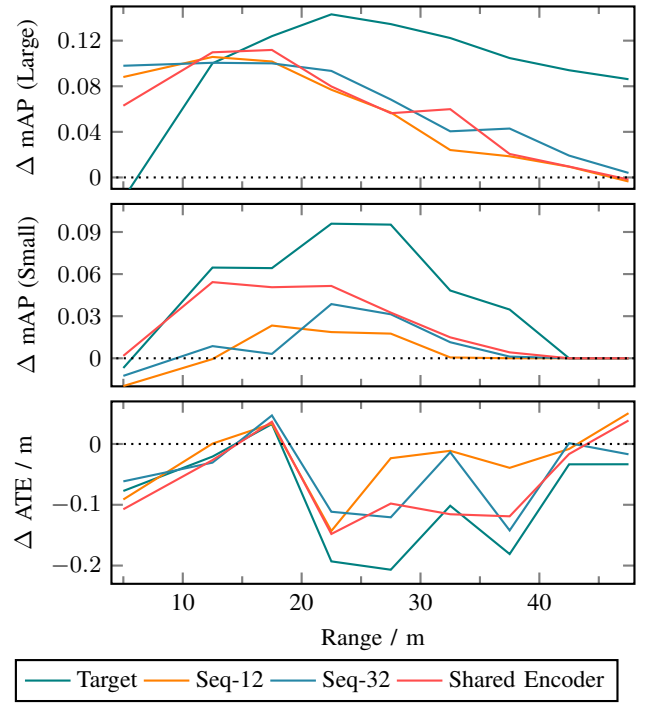
\begin{figure}[ht]
    \centering
    \begin{tikzpicture}
        \tikzstyle{every node}=[font=\small]
        \begin{groupplot}[group style={group size=1 by 3, vertical sep=2mm}]
            \nextgroupplot[
                xmin=4,
                xmax=48,
                ymin=-0.01,
                ymax=0.15,
                xticklabels={,,},
                minor x tick num=1,
                ytick={0, 0.04, 0.08, 0.12},
                yticklabel style={/pgf/number format/fixed, /pgf/number format/precision=2},
                ylabel={$\Delta$ mAP (Large)},
                legend columns=4,
                legend to name=legendmetricsvsdistance,
            ]
            \addplot+[mark=None] table [x=Distance, y expr=\thisrow{Target}-\thisrow{Baseline}] from \mAPLarge;
            \addlegendentry{Target};
            \addplot+[mark=None] table [x=Distance, y expr=\thisrow{Seq12}-\thisrow{Baseline}] from \mAPLarge;
            \addlegendentry{Seq-12};
            \addplot+[mark=None] table [x=Distance, y expr=\thisrow{Seq32}-\thisrow{Baseline}] from \mAPLarge;
            \addlegendentry{Seq-32};
            \addplot+[mark=None] table [x=Distance, y expr=\thisrow{SE}-\thisrow{Baseline}] from \mAPLarge;
            \addlegendentry{Shared Encoder};
            \draw [dotted] (0,0) -- (50,0);

            \nextgroupplot[
                xmin=4,
                xmax=48,
                ymin=-0.02,
                ymax=0.11,
                xticklabels={,,},
                minor x tick num=1,
                ytick={0, 0.03, 0.06, 0.09},
                yticklabel style={/pgf/number format/fixed, /pgf/number format/precision=3},
                scaled y ticks=false,
                ylabel={$\Delta$ mAP (Small)},
            ]
            \addplot+[mark=None] table [x=Distance, y expr=\thisrow{Target}-\thisrow{Baseline}] from \mAPSmall;
            \addplot+[mark=None] table [x=Distance, y expr=\thisrow{Seq12}-\thisrow{Baseline}] from \mAPSmall;
            \addplot+[mark=None] table [x=Distance, y expr=\thisrow{Seq32}-\thisrow{Baseline}] from \mAPSmall;
            \addplot+[mark=None] table [x=Distance, y expr=\thisrow{SE}-\thisrow{Baseline}] from \mAPSmall;
            \draw [dotted] (0,0) -- (50,0);

            \nextgroupplot[
                xmin=4,
                xmax=48,
                ymin=-0.23,
                ymax=0.07,
                xlabel={Range / m},
                xticklabels={-10, 0, 10, 20, 30, 40, 50},
                minor x tick num=1,
                ytick={-0.2, -0.1, 0, 0.1},
                yticklabel style={/pgf/number format/precision=3},
                ylabel={$\Delta$ ATE / m},
                scaled y ticks=false,
            ]
            \addplot+[mark=None] table [x=Distance, y expr=\thisrow{Target}-\thisrow{Baseline}] from \ATE;
            \addplot+[mark=None] table [x=Distance, y expr=\thisrow{Seq12}-\thisrow{Baseline}] from \ATE;
            \addplot+[mark=None] table [x=Distance, y expr=\thisrow{Seq32}-\thisrow{Baseline}] from \ATE;
            \addplot+[mark=None] table [x=Distance, y expr=\thisrow{SE}-\thisrow{Baseline}] from \ATE;
            \draw [dotted] (0,0) -- (50,0);
        \end{groupplot}
    \end{tikzpicture}
    \begin{small}
        \ref{legendmetricsvsdistance}
    \end{small}
    \caption{
        Relative improvement (mAP and ATE) of our models compared to the baseline model along the detection range.
        mAP values $>0$ and ATE values $<0$ denote an improvement.
    } \label{fig:metrics_vs_distance}
\end{figure}

\Cref{tab:od_quanti} summarizes average precision (AP) for large and small objects, mean average translation error (ATE) and mean average scale error (ASE) for all experiments.
We compare our models against a \textit{Baseline} object detector which uses the single frame layer in \cref{tab:used_layers} as input.
In contrast, the \textit{Target} configuration uses all fused layers as input and may denote an upper performance boundary for object detection as all layers are used.

We observe that although our models do not improve the AP of frequent large objects such as cars and trucks, the AP of small objects is improved for almost all models compared to the baseline.
This leads to a higher mAP value than the baseline for our models.
Here, the shared encoder model achieves the best AP for pedestrians, motorcycles and bicycles.
We also see that the additional enrichment task increases the localization and scale accuracy (lower mATE and mASE compared to the baseline).

\Cref{fig:qualitative_results} compares the object detection result between \textit{Baseline} and \textit{SE} for one particular scenario.
The shared encoder model is able to detect also the group of pedestrians on the sidewalk. 

As illustrated in \cref{fig:metrics_vs_distance}, we observe that an additional enrichment improves the detection over a long range in most cases.
However, the performance deteriorates slightly for close distances and for localization at certain ranges.
We believe this is due to the abundance of well-visible object these distances. \section{Conclusion} \label{sec:conclusion}

We are able to improve object detection performance for small objects and objects at larger distances by incorporating structural knowledge in terms of an enriched map into the training process.
Additionally, the resulting multi-task model can be used to generate enriched versions of single-frame measurement inputs which may be used for subsequent methods such as state estimation via particle filters.
Therefore we presented an evidential fusion approach based on known poses which assigns high uncertainty to contradicting measurement, e.g. in the presence of moving traffic participants.
This training data can be generated in an automatic fashion and does not require additional manual annotations.
As the beliefs were estimated conservatively with many measurements assigned to uncertainty, future work is to estimate the grid cell state in an acausal manner to further improve the target data quality. 
\printbibliography

\end{document}